\documentclass[sigconf,screen]{acmart}
\usepackage{multirow}
\usepackage{subfigure}
\usepackage{blindtext}
\usepackage{enumitem}
\usepackage{algorithm}  
\usepackage{algpseudocode}  
\usepackage{amsmath}  
\usepackage{hyperref}
\usepackage{flushend}
\usepackage{balance}
\usepackage{mathtools}
\usepackage{bbm}
\DeclareMathOperator*{\maximize}{\text{\fontfamily{pcr}\selectfont maximize}}
\AtBeginDocument{%
\providecommand\BibTeX{{%
  \normalfont B\kern-0.5em{\scshape i\kern-0.25em b}\kern-0.8em\TeX}}}

  \copyrightyear{2023} 
  \acmYear{2023} 
  \setcopyright{acmcopyright}
  \acmConference[RecSys'23]{the International Workshop on Deep Learning Practice for High-Dimensional Sparse Data}{Sept 18--22, 2023}{Singapore}
  \acmBooktitle{the International Workshop on Deep Learning Practice for High-Dimensional Sparse Data (DLP@RecSys '23), Sept 18--22, 2023, Singapore}
  \acmPrice{15.00}
  \acmDOI{10.1145/xxxxxx.xxxxxx}
  \acmISBN{978-1-4503-xxxx-x/xx/xx}

\begin{document}

\title{NMA: Neural Multi-slot Auctions with Externalities for Online Advertising}

\author{Guogang Liao}
\authornote{Equal contribution. Listing order is random.}
\affiliation{%
 \institution{Meituan}
 \city{Beijing}
 \country{China}
}
\email{liaoguogang@meituan.com}

\author{Xuejian Li}
\authornotemark[1]
\affiliation{%
 \institution{Meituan}
 \city{Beijing}
 \country{China}
}
\email{lixuejian03@meituan.com}

\author{Ze Wang}
\authornotemark[1]
\authornote{Corresponding author.}
\affiliation{%
 \institution{Meituan}
 \city{Beijing}
 \country{China}
}
\email{wangze18@meituan.com}

\author{Fan Yang}
\authornotemark[1]
\affiliation{%
 \institution{Meituan}
 \city{Beijing}
 \country{China}
}
\email{yangfan129@meituan.com}

\author{Muzhi Guan}
\affiliation{%
 \institution{Meituan}
 \city{Beijing}
 \country{China}
}
\email{guanmuzhi@meituan.com}

\author{Bingqi Zhu}
\affiliation{%
 \institution{Meituan}
 \city{Beijing}
 \country{China}
}
\email{zhubingqi@meituan.com}

\author{Yongkang Wang}
\affiliation{%
 \institution{Meituan}
 \city{Beijing}
 \country{China}
}
\email{wangyongkang03@meituan.com}

\author{Xingxing Wang}
\affiliation{%
 \institution{Meituan}
 \city{Beijing}
 \country{China}
}
\email{wangxingxing04@meituan.com}

\author{Dong Wang}
\affiliation{%
 \institution{Meituan}
 \city{Beijing}
 \country{China}
}
\email{wangdong07@meituan.com}
\renewcommand{\shortauthors}{Guogang Liao, et al.}

\begin{abstract}
  Online advertising driven by auctions brings billions of dollars in revenue for social networking services and e-commerce platforms. 
  GSP auctions, which are simple and easy to understand for advertisers, have almost become the benchmark for ad auction mechanisms in the industry. 
  However, most GSP-based industrial practices assume that the user click only relies on the ad itself, which overlook the effect of external items, referred to as externalities. 
  Recently, DNA has attempted to upgrade GSP with deep neural networks and models local externalities to some extent. However, it only considers set-level contexts from auctions and ignores the order and displayed position of ads, which is still suboptimal.
  Although VCG-based multi-slot auctions (e.g., VCG, WVCG) make it theoretically possible to model global externalities (e.g., the order and positions of ads and so on), they lack an efficient balance of both revenue and social welfare.
    
  In this paper, we propose novel auction mechanisms named \textit{Neural Multi-slot Auctions} (NMA) to tackle the above-mentioned challenges. Specifically, we model the global externalities effectively with a context-aware list-wise prediction module to achieve better performance. We design a list-wise deep rank module to guarantee incentive compatibility in end-to-end learning. Furthermore, we propose an auxiliary loss for social welfare to effectively reduce the decline of social welfare while maximizing revenue. 
    Experiment results on both offline large-scale datasets and online A/B tests demonstrate that NMA obtains higher revenue with balanced social welfare than other existing auction mechanisms (i.e., GSP, DNA, WVCG) in industrial practice, and we have successfully deployed NMA on Meituan food delivery platform.

\end{abstract}



\maketitle

\section{Introduction}
\label{sec:1}
\begin{figure}[tb]
  \centering
  \includegraphics[width=0.7\linewidth]{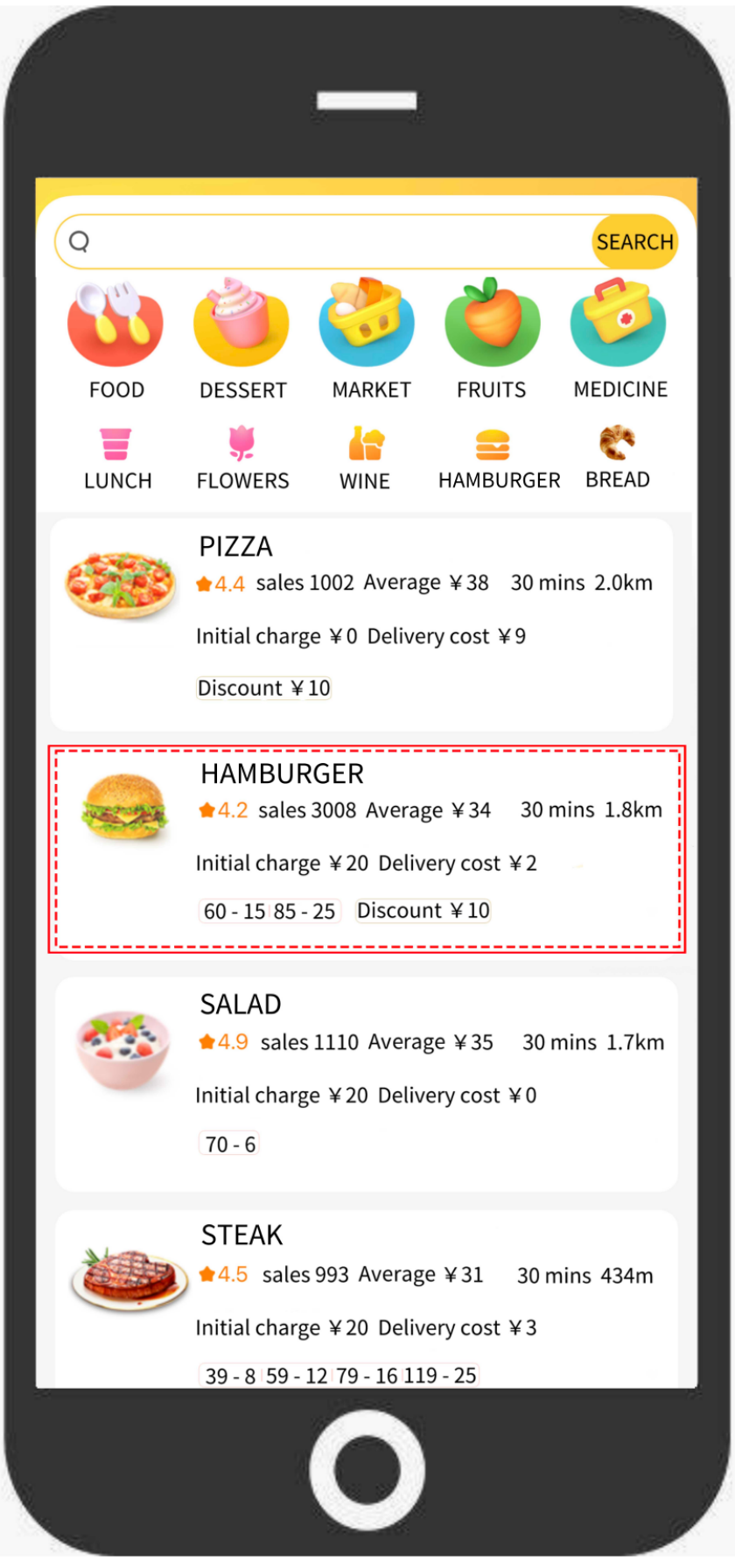}
  \caption{
    On Meituan food delivery platform, the ad 'Hamburger' is presented to user along with external items. Whether a user clicks on the ad is easily influenced by both the position of the ad and the context of the ad.
  }
  \label{fig:fig_intro}
\end{figure}

Nowadays, many online platforms such as Google, Facebook, Baidu, Alibaba, etc, organize their contents in feeds \cite{chen2022hierarchically,habib2022identifying, liu2022lion, ma2022online}. Feeds provide advertisers with a huge traffic platform, making online advertising the main source of income for these companies. 
Unlike organic items only ranked by user preference \cite{wang2022surrogate,zheng2022disentangling, jin2022learn}, the display of ads depends on user preference, platform revenue and advertiser's utility \cite{aggarwal2022simple, zibin2023adverse, despotakis2021first}. Accordingly, it is essential for online platforms to design proper auction mechanisms, which benefits the users, the advertisers and the platform at the same time.

% GSP
\textbf{Generalized Second Price (GSP)} auctions \cite{edelman2007internet}, which has nice interpretation and is easy to deploy in industry, has almost become the benchmark for ad auction mechanisms.
However, directly applying GSP to our platform faces one major problem: GSP assumes that the user click only relies on ad itself \cite{edelman2007internet, zhang2021optimizing, liu2021neural}, which is difficult to meet in practice. As shown in Figure \ref{fig:fig_intro}, the ad 'Hamburger' is presented to user along with external items in our scenario. As a result, whether a user clicks on the ad is easily influenced by both the position of the ad and the context of the ad, which violates the assumption of GSP. 
% DNA
Recently, \citet{liu2021neural} propose the GSP-based \textbf{Deep Neural Auctions (DNA)} \cite{liu2021neural}, which exploit the set-level information of auction environment and leverage modern machine learning to enlarge the design space of auction mechanisms. However, DNA only model local externalities and ignore the important information about the order and displayed position of ads, which are still suboptimal. 
% VCG
Compared with GSP and DNA, the representative truth-telling auctions \textbf{Vickrey-Clarke-Groves (VCG)} \cite{varian2014vcg, varian2007position, gomes2014bayes} take the impact of external items into account and  make it theoretically possible to model global externalities. 
However, VCG causes the decline of platform revenue, which is unacceptable for most industrial practice. 
% WVCG
\textbf{Weighted Vickrey-Clarke-Groves (WVCG)} \cite{gatti2015truthful} is proposed to model the global externalities and solve the revenue decline problem of VCG to some extent. However, the high complexity of parameter solving makes it difficult to practically apply in industrial scenario.

By analyzing the problems of these representative auction mechanisms mentioned above, we consider three critical challenges on designing multi-slot auction mechanisms with externalities in industry: 
i) Most auction mechanisms either only model local externalities or model global externalities inefficiently. How to model global externalities efficiently has become a key point. 
ii) Both platform revenue and social welfare(SW)\cite{wang2022designing} are the core metrics on designing auction mechanisms. Most existing works focus on the maximization of either platform revenue or SW, lacking an effective balance between the two.
iii) Some methods focus on the theoretical auction settings, in lack of the insights from large-scale industrial deployment.

To this end, we design end-to-end learning multi-slot auction mechanisms with externalities for online advertising, named \textit{\textbf{N}eural \textbf{M}ulti-slot \textbf{A}uctions} (NMA), to maximize platform revenue with less SW decline. Specifically, the structure of NMA consists of four parts:
Firstly, all candidate allocations\footnote{Through full permutation algorithm, all allocations are generated in the form of ad list. In industrial practice, the full permutation algorithm can be replaced by a list generation module, such as heuristic list generation or model-based method like Seq2slate,PIER. } are put into a context-aware list-wise prediction module, which effectively models the global externalities.
Secondly, we propose a list-wise deep rank module by modeling the parameters as a neural network, which makes it possible to train the complex auction mechanisms end-to-end. 
Thirdly, we design a list-wise differentiable sorting module to effectively train the cascaded deep neural networks end-to-end by real system reward feedback. 
Finally, We further design an auxiliary loss for SW to effectively reduce its decline while maximizing revenue.
The main contributions are summarized as follows: 
\begin{itemize}[leftmargin=*]
 \item \textbf{A Novel framework for multi-slot auctions.} In this paper, we propose a novel end-to-end learning framework for multi-slot auctions named NMA, which enables accurate modeling of global externalities and effective balancing of key indicators on designing auction mechanisms.
 \item \textbf{A superior approach to modeling global externalities.} We employ a finely designed neural network to model the list-level information of each allocation and public information in auctions, which greatly improves the effectiveness of global externality modeling and the accuracy of predictive values.
 \item \textbf{An efficient method of balancing revenue and SW.} We design an auxiliary loss for social welfare maximization, which is able to effectively balance platform revenue and social welfare.
 \item \textbf{Detailed industrial and practical experience.} We successfully deploy NMA on Meituan food delivery platform. The results of both offline simulation and online A/B experiment demonstrate that NMA brings a significant improvement in CTR and platform revenue, compared with GSP \cite{edelman2007internet}, DNA \cite{liu2021neural}, VCG \cite{varian2014vcg} and WVCG \cite{gatti2015truthful}.
 \end{itemize} 

 \section{related work}
 The externalities modeling \cite{gomes2009externalities, gatti2012truthful, hummel2014position, gatti2015truthful} in this paper refers to considering both position-dependent externalities and ad-dependent externalities. In other words,
 whether a user clicks on an ad depends not only relies on the ad itself, but also on where the ad is placed and the context in auctions.
 Most classical auction mechanisms such as GSP \cite{edelman2007internet} and uGSP \cite{bachrach2014optimising} auctions assume that the user click only relies on ad itself. Therefore, they lack modeling of externalities, resulting in suboptimal performance.
 
 \textbf{Local Externalities Modeling}. Recently, academia researchers and industry practitioners are becoming increasing aware of the importance of externalities modeling and have tried to model local externalities \cite{liu2021neural, huang2021deep, ghosh2008externalities,fotakis2011externalities}. 
 For instance, Deep GSP \cite{zhang2021optimizing} and DNA \cite{liu2021neural} proposed by \emph{Alibaba} models the auction mechanisms as deep neural networks, where the feedback from real system is the reward for end-to-end learning. The set-level information of candidate ads used in DNA can be regarded as the consideration of local ad-dependent externalities. However, DNA ignores the order and displayed position of ads and is still suboptimal.
 Besides, some researchers design an additional deep neural networks to model externalities and complete allocation and payment based on classical auction mechanisms such as GSP and VCG.
 For example, \citet{huang2021deep} propose Deep Position-wise Interaction Network (DPIN) to model position-dependent externalities on \emph{Meituan}, which predicts the CTR of each ad in each position and divides the molti-slots auctions into multiple rounds of single-slot auction based on GSP. However, DPIN does not consider ad-dependent externalities. 
 User browsing model (UBM) \cite{dupret2008user} applied in \emph{Baidu} and cascade models \cite{kempe2008cascade, fotakis2011externalities, farina2016ad} take position-dependent externalities and ad-dependent externalities into consideration. But they only model the impact of the preceding ads on the following ads, ignoring the impact of the following ads on the preceding ads.
 
 \textbf{Global Externalities Modeling}.
 VCG\cite{varian2014vcg} evaluates all candidate allocations by social welfare and provides the possibility of global externalities modeling. 
 In VCG, the allocation with the largest social welfare wins and each ad in the winning allocation is charged for the decline of social welfare caused by the ad.
 Compared with GSP, VCG leads to lower platform revenue, which hinders its application in industry \cite{varian2007position}. WVCG with cascade model \cite{gatti2015truthful} is proposed to solve these problems. In WVCG, the social welfare of each allocation is linearly weighted by parameters to achieve higher revenue. However, \citet{jeziorski2015makes} find that the cascade models fail to model global externalities when user browse other than top-to-bottom. Furthermore, WVCG solves parameters by Multi-Armed Bandit (MAB), which has high complexity of parameter solving and makes it difficult to practice in industrial scenario. 
 Recently, Automated Mechanism Design (AMD) methods such as Virtual Valuations Combinatorial auctions (VVCA) and Affine Maximizer Auctions (AMA) \cite{sandholm2015automated} are proposed to improve the auctioneer income by affine transformation of social welfare. But they have the same problem as WVCG in parameter solving and the decline of social welfare. Therefore, we use representative VCG and WVCG for comparison and no longer discuss other similar methods such as VVCA and AMA.

 Notice that, some auction mechanisms for online advertising such as two-stage auction \cite{wang2022designing}, DPIN \cite{huang2021deep}, UBM \cite{dupret2008user} and cascade models \cite{kempe2008cascade, fotakis2011externalities, farina2016ad}, are not discussed and compared in this paper due to different application scenarios.

 \section{Preliminaries}

\subsection{Setting for Multi-slot Auctions}
We describe a typical setting for multi-slot auctions in e-commerce advertising. 
Formally, when a user initiates a page view request, the e-commerce platform displays $J$ items to the user\footnote{We assume that the $J$ items must be displayed to simplify the problem.}, which contains $K (K \leq J)$ ads. $N$ advertisers compete for the $K (K \leq N)$ ad slots and each advertiser\footnote{Each advertiser has only one ad in our scenario and therefore we can identify by $a_i$ the i-th ad the the i-th advertiser indifferently.} $a_i$  submits a bid $b_i$ for a slot according to private information (e.g., $pCTR$, etc). We present the auction mechanisms by $\mathcal{M}\langle \mathcal{R}, \mathcal{P} \rangle$. $\mathcal{R}$ is the ad allocation scheme, which is used to select $K$ winning ads from $N$ candidate advertisers and display them on the corresponding $K$ fixed ad slots from top to bottom in this request. $\mathcal{P}$ is the payment rule, which is used to calculate the payments for winning ads and would be carefully designed to guarantee the economic properties and the revenue of the auction mechanisms.

\subsection{Problem Formulation}
In multi-slot auctions, we denote the set of ads as $\mathcal{A} = \{a_1,\dots,a_N\}$, the set of ad slots as $\mathcal{K} = \{1,\dots,K\}$. Then an allocation $\theta$ is defined as selecting $K$ ads from $N$ candidates and placing them on $K$ slots in order.
Given bids from advertisers and predicted CTRs, we aim to design multi-slot auction mechanisms with externalities $\mathcal{M}\langle \mathcal{R}, \mathcal{P} \rangle$ to maximize platform revenue and reduce the decline of social welfare under basic assumptions, as follows:
\begin{equation}
  \begin{aligned}
  \maximize_{\mathcal{M}} \quad & \mathbb{E}_{\mathbf{\theta} \in \Theta} [\sum_{a_j \in \text{ads}(\theta)} p(\theta,a_j) \cdot \hat q(\theta,a_j)],\\
  \textrm{s.t.} \quad 
  & \textit{Incentive Compatibility (IC) constraint,}\\
  & \textit{Individual Rationality (IR) constraint,}\\
  & \textit{Social Welfare (SW) constraint,}\\
  \end{aligned}
  \label{eq:problem}
\end{equation}
where $\Theta$ is the set of all possible allocations, $\theta^*$ is the best allocation, $\text{ads}(\theta)$ is a subset of ads allocated in $\theta$, $p(\theta,a_j)$ is the payment of $a_j$ in $\theta$, and $\hat q(\theta,a_j)$ is the predicted CTR of $a_j$ in $\theta$.

The constraints of IC and IR \cite{edelman2007internet,liu2021neural,varian2014vcg,gatti2015truthful} guarantee that advertisers would truthfully report the bid, and would not be charged more than their maximum willing-to-pay for the allocation \cite{qin2015sponsored,liu2021neural}. 

Social welfare is a crucial metric for online advertising, as it measures the efficiency on matching advertisers and users, and is also the upper bound of the revenue which is the sum of the total payments of the ad platform \cite{wang2022designing}. The constraint of SW is defined as follows:
\begin{equation}
  \label{eq:swc}
  1 - \frac{SW(\theta^*, \textbf{b})}{SW^*} < \varepsilon,
\end{equation}
where $SW(\theta^*, \textbf{b})$ is the SW of the best allocation $\theta^*$, $SW^*$ is the maximum SW in $\Theta$, and $\varepsilon$ is the decline threshold of SW formulated according to the business scenario.

The allocation scheme and payment rule in $\mathcal{M}\langle \mathcal{R}, \mathcal{P} \rangle$ are:

$\bullet$ Allocation scheme $\mathcal{R}:\times_{a_j \in \mathcal{A}} \mathcal{V}_{j}\longrightarrow \Theta$.

$\bullet$ Payment rule for each ad $\mathcal{P}:\times_{a_j \in \mathcal{A}} \mathcal{B}_{j}\longrightarrow \mathbb{R}^{+}$,

\noindent where $\mathcal{V}_j$ is the set of possible values for ad $a_j$ and $\mathcal{B}_j$ is the set of possible bids for ad $a_j$. Specifically, we denote the real click value of $a_i$ as $v_i$, the submitted click value of $a_j$ as $b_j$.

An auction mechanism is IC if it is in the best interest of each advertiser to truthfully reveal her maximum willing-to-pay price, i.e, $b_j=v_j$. An auction mechanism is IR if the payment of advertiser $a_j$ would not exceed the reported value, i.e., $p(\theta,a_j)<b_j$ if ad $a_j$ is displayed and clicked; or pays nothing, otherwise. With these two properties, advertisers do not need to spend efforts in computing bidding strategy, and are encouraged to participate in the auctions with no risk of deficit. The online platform also obtains the truthful and reliable advertisers' values.
Since we follow the theoretical basis of WVCG which is proven IC and IR \cite{gatti2015truthful}, we identify by $b_i$ the real click value of the $i$-th advertiser.

We list the key symbols used in this paper in Table \ref{table:notation} for clarity.

\begin{table}[tb]
  \vspace{0.05in}
  \centering
  \caption{List of key symbols.}
  \setlength{\tabcolsep}{3pt}
  \begin{tabular}{c|c}
    \hline
    Symbol & Meaning \\
    \hline
    $N$ & Number of ads  \\
    $K, K \leq N$ & Number of available ad slots  \\
    $N_{L}$ & Number of candidate ad allocations  \\
    $\mathcal{A} = \{a_1,\dots,a_N\}$ & Set of candidate ads \\
    $\mathcal{K} = \{1,\dots,K\}$ & Set of ad slots \\
    $\mathcal{B}_j = [0,B_j], B_j \in  \mathbb{R}^{+}$ & Set of the possible bids for $a_j$\\
    $v_{j}\in \mathcal{V}_j$ & Real click value of $a_j$ \\
    $b_{j}\in \mathcal{B}_j$ & Submitted click value (i.e., bid) of $a_j$ \\
    $\textbf{v}=(v_1,\dots,v_N)$ & Values profile of ads \\
    $\textbf{b}=(b_1,\dots,b_N)$ & Bids profile of ads  \\
    $\Theta=\{\theta_1,\dots,\theta_{N_{L}}\}$ & Set of all possible allocations \\
    $\theta \in \Theta$ & An allocation (i.e., an ordered ad list) \\
    $\theta^* \in \Theta$ & The best allocation \\
    $\theta_{-j} \in \Theta$ & An allocation when $a_j$ is not present  \\
    $\theta^*_{-j} \in \Theta$ & The best allocation when $a_j$ is not present \\
    $\text{ads}(\theta) \subseteq \mathcal{A}$ & Subset of ads allocated in $\theta$ \\

    $\hat q(\theta,a_j)$ & Predicted CTR of $a_j$ in $\theta$ \\
    $p(\theta,a_j)$ & Payment of $a_j$ in $\theta$ \\
    $RS(\theta, \textbf{b})$ &  Ranking score (RS) of $\theta$ \\
    $RS_{-j}(\theta, \textbf{b})$ & RS of $\theta$ minus RS of $a_{j}$ \\
    $SW(\theta, \textbf{b})$ & Social welfare of $\theta$ \\
    \hline
  \end{tabular}
  \vspace{0.05in}
  \label{table:notation}
  \vspace{-0.15in}
\end{table}

\section{Methodology}
\begin{figure*}[tb]
  \centering
  \includegraphics[width=\textwidth]{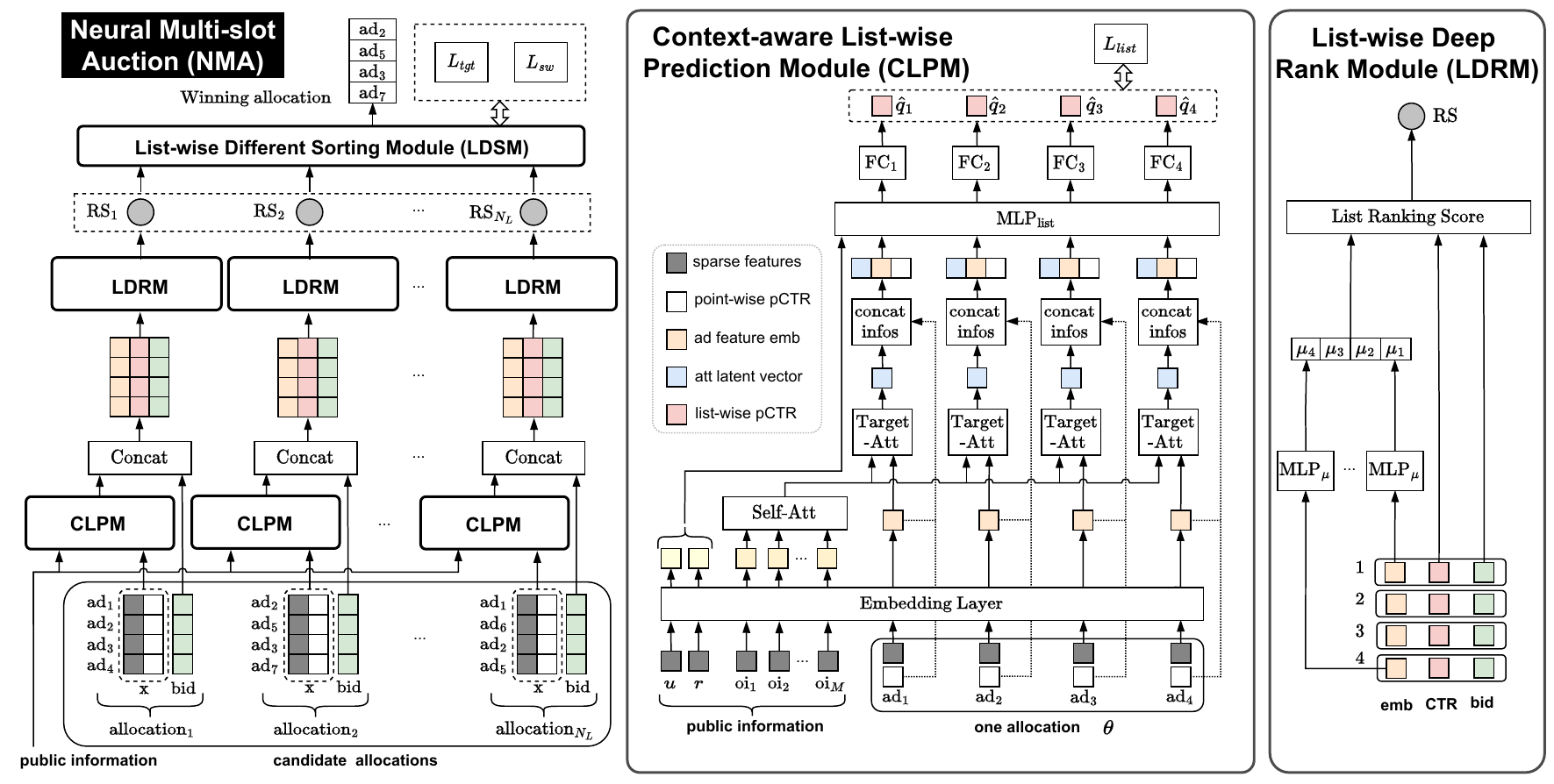}

  \caption{
  The Network architecture of NMA. NMA takes all candidate ad lists and  three types of public information as input and uses three novel modules to select the best ad list. The three modules of NMA are the Context-aware List-wise Prediction Module (CLPM), the List-wise Deep Rank Module (LDRM) and List-wise Differentiable Sorting Module (LDSM), respectively.
  }
  \label{fig:dca}
\end{figure*}

In this section, we present the details of Neural Multi-slot Auctions (NMA) with externalities, which maximizes platform revenue with less decline of social welfare for online advertising. As shown in Figure \ref{fig:dca}, NMA takes all candidate ad lists (i.e., each ad list represents an allocation $\theta$) and three types of public information as input, and uses three novel modules to select the winning ad list. 
The three modules of NMA are the Context-aware List-wise Prediction Module (CLPM), the List-wise Deep Rank Module (LDRM) and the List-wise Differentiable Sorting Module (LDSM). Specifically, CLPM models both position-dependent externality and ad-dependent externality of each ad list and predicts the list-wise pCTR of each ad in the ad list based on the public information of the request and the contextual information of the ad list. LDRM combines auction mechanisms with deep learning together and uses a  sub-network to model the information of each ad list, with the aim of improving the expressive ability of ranking formula while guaranteeing IC and IR properties. LDSM conducts a continuous relaxation of sorting operator in auctions, and outputs a row-stochastic permutation matrix. We can use this row-stochastic permutation matrix to express the expected revenue and learn the optimal list end-to-end. Meanwhile, we propose a social welfare maximization auxiliary loss to meet the social welfare constraints and reduce the decline of social welfare. 
Next we will introduce each part separately.

\subsection{Context-aware List-wise Prediction Module}
\label{sec:clpm}
Most traditional advertising auction mechanisms use point-wise pCTR\footnote{The point-wise pCTR means CTR predicted based on the information of user-ad pair, without considering the impact of displayed position and auction environment.} to calculate the ranking score for the process of allocation and payment. However, the point-wise pCTR not only lacks the utilization of public information in auctions, but also fails to consider the interaction between ads in one list \cite{pei2019personalized,carbonell1998use,zhai2015beyond}. 
Recently \citet{liu2021neural} take the interaction of ads into consideration and propose a module named set-encoder to automatically model the set-level information in auctions. But it only models local externalities and thus still suboptimal. 
To this end, we propose the context-aware list-wise prediction module to model global externalities explicitly, outputting the more accurate list-wise pCTR of each ad in the list. 

CLPM adopts a parameter-sharing structure for all candidate ad lists. Here we take an ad list $\theta$ as an example to illustrate. As shown in Figure \ref{fig:dca}, CLPM takes the ad list $\theta$ and three types of public information (i.e., request information, user profile and organic items \cite{liao2021cross,yan2020LinkedInGEA} in current request) as input and outputs the  list-wise pCTR of each ad in the list. We first use embedding layers to extract the embeddings from raw inputs. The embedding matrixes for the ad list and organic items list are denoted as $\mathbf{E}_{\text{ad}} \in \mathbb{R}^{K \times d}$ and $\mathbf{E}_{\text{oi}} \in \mathbb{R}^{M \times d}$,  where $K$ is the number of ads in $\theta$, $M$ is the number of organic items in current request and $d$ is the dimension of embedding.
Meanwhile, the embeddings for user profile and request information are denoted as $\mathbf{e}^\text{u}$ and $\mathbf{e}^\text{r}$. 

Then we use a Self Attention Unit \cite{vaswani2017attention} to model the sequence information of organic items list:
\begin{equation}
  \begin{aligned}
    \mathbf{H}_\text{oi} = \text{Self-Att}(\mathbf{Q}_\text{oi},\mathbf{K}_\text{oi},\mathbf{V}_\text{oi}) = \text{soft} \max (\frac{\mathbf{Q}_\text{oi}\mathbf{K}_\text{oi}^\top}{\sqrt{d}})\mathbf{V}_\text{oi},
  \end{aligned}
\end{equation}
where $\mathbf{Q}_\text{oi},\mathbf{K}_\text{oi},\mathbf{V}_\text{oi}$ represent query, key, and value, respectively. Here query, key, and value are transformed linearly from embedding matrix of organic items, as follows:
\begin{equation}
  \begin{aligned}
    \mathbf{Q}_\text{oi} = \mathbf{E}_\text{oi}\mathbf{W}^{Q}_\text{oi}, \mathbf{K}_\text{oi} = \mathbf{E}_\text{oi}\mathbf{W}^{K}_\text{oi},\mathbf{V}_\text{oi} = \mathbf{E}_\text{oi} \mathbf{W}^{V}_\text{oi}.
  \end{aligned}
\end{equation}
    
Next, we use a Target Attention Unit \cite{vaswani2017attention,zhou2018DIN} to encode the interaction between the organic items and each ad in the ad list:
\begin{equation}
\begin{aligned}
  \mathbf{h}^\text{ad}_j&= \text{Tgt-Att} \Big( \mathbf{e}^\text{ad}_{j}, \{\mathbf{h}^\text{oi}_{i} \}_{i=1}^{M} \Big) \\
    & =\mathbf{e}^\text{ad}_{j} \! \cdot \!\text{MLP}_{\text{att}} (\mathbf{e}^\text{ad}_{j}|\mathbf{h}^\text{oi}_{1}) + \cdots + \mathbf{e}^\text{ad}_{j}\! \cdot\! \text{MLP}_{\text{att}} (\mathbf{e}^\text{ad}_{j}|\mathbf{h}^\text{oi}_{M}) , \forall j\in [K],
\end{aligned}
\end{equation}
where $|$ means concatenation, $\mathbf{e}^\text{ad}_{j}$ is the embedding of the $j$-th ad in the ad list $\theta$, $\text{MLP}_{\text{att}}$ is a Multi-Layer Perceptron (MLP) which takes embeddings of ad and organic item pair as input and outputs an attention weight, and $\mathbf{h}^\text{oi}_{i}$ is the representation of the $i$-th organic item in the organic items list generated from previous unit. 

Then we concatenate the representation of the $j$-th ad with its corresponding embedding and point-wise pCTR together as its final representation. All representations of ads in the ad list $\theta$ are input into an MLP to model the global externalities:
\begin{equation}
  \begin{aligned}
    \mathbf{e}^\text{list}= \text{MLP}_{\text{list}} \Big( (\mathbf{e}^\text{ad}_{1}|\mathbf{h}^\text{ad}_{1}|\text{CTR}^\text{ad}_{1})|\cdots|(\mathbf{e}^\text{ad}_{K}|\mathbf{h}^\text{ad}_{K}|\text{CTR}^\text{ad}_{K}) \Big),
\end{aligned}
\end{equation}
\begin{equation}
  \begin{aligned}
    \hat q(\theta,a_j)= \sigma \Big( \text{FC}_j ( \mathbf{e}^\text{list} | \mathbf{e}^u  | \mathbf{e}^r)\Big), \forall j\in [K],
\end{aligned}
\end{equation}
where $\text{CTR}$ denotes the point-wise pCTR, $\sigma$ denotes the sigmoid function and $\hat q(\theta,a_j)$ denotes the list-wise pCTR of the $j$-th in the ad list $\theta$. In order to make the CLPM converge better, we propose an auxiliary loss for CLPM based on the real feedback of each ad:
\begin{equation}
  \begin{aligned}
    L_{list} = \sum_{j=1}^{K}\Big(-y_j^{\text{ad}}\log ( \hat q(\theta,a_j)) - (1-y_j^{\text{ad}})\log(1-\hat q(\theta,a_j))\Big),
  \end{aligned}
  \end{equation}
where $y_j^{\text{ad}} \in \{0,1\}$ represents whether user clicks the $j$-th ad in the ad list $\theta$ or not. 
Through CLPM, we can effectively model the global externalities, outputting the list-wise pCTRs of the ad list. Compared with point-wise pCTR, the list-wise pCTR is able to model the position-dependent externality and is more accurate, which can help NMA achieve better performance.
Notice that, here we only take CTR as an example. The framework of CLPM, which improves the accuracy of predicted values through modeling global externalities, can be easily transferred to other prediction problems, such as for CVR prediction, order price prediction, etc.

\subsection{List-wise Deep Rank Module}
\label{sec:ldrm}
In LDRM, we combine auction mechanisms with deep learning together to improve the expressive ability of ranking formula while guaranteeing IC and IR properties.
First we extract bids, embeddings, the list-wise pCTRs of ads in the ad list $\theta$ as input. Following the auction theory \cite{gatti2015truthful, sandholm2015automated}, we design a sub-network (i.e., $\mu$-Net) to calculate the ranking score of each ad list. The output of the $\mu$-Net corresponds to the weight in WVCG, which is used as a weighting factor to boost revenue. However, the MAB method of WVCG is difficult to optimize in high-industrial scenarios. Therefore, we upgraded the deep neural network for end-to-end optimization, which have rich expressive ability and can automatically optimize affine parameters. 

Mathematically, we formulate $\mu$-Net as $f_{\mu}(\cdot)$. which takes the allocation-independent features of each ad as input to ensure IC and outputs the private value of corresponding ad\cite{gatti2015truthful}:
\begin{equation}
  \begin{aligned}
    f_{\mu}(a_j)= \sigma \Big( \text{MLP}_\mu (\mathbf{e}^{\text{ad}}_j) \Big), \forall j\in [K].
  \end{aligned}
\end{equation}

Then the ranking score of the ad list $\theta$ is calculated as \cite{sandholm2015automated}:
\begin{equation}
  \begin{aligned}
    RS(\theta, \textbf{b}) &= \sum_{a_j \in \text{ads}(\theta)} f_{\mu}(a_j) \cdot b_{j} \cdot  \hat q(\theta,a_j) \\
&=  \sum_{j=1}^K f_{\mu}(a_j) \cdot {b}_j \cdot  \hat q(\theta,a_j).
  \end{aligned}
\end{equation}

The payment rule of the best ad list $\theta^{*}$ is:
  \begin{equation}
    \begin{aligned}
    p(\theta,a_j)\! =\! \frac{1}{f_{\mu}(a_{j})\! \cdot\! \hat q(\theta,a_j)}[RS(\theta^{*}_{\!-j}, \textbf{b}_{\!-j})\!-\!RS_{-j}(\theta^{*}, \textbf{b})], \forall j\!\in\! [K],
     \label{eq:pay}
     \end{aligned}
  \end{equation}
where $a_{j} \in \text{ads}(\theta^{*})$ and $\theta^{*}_{-j}$ is the best ad list when $a_{j}$ is not present. 

Obviously given the payment rule, the payment of each ad is lower than its bid \cite{sandholm2015automated}, which guarantees IR of NMA.
The expected revenue of the ad list $\theta$ is calculated as:
\begin{equation}
  \begin{aligned}
   r(\theta) =  \sum_{j=1}^K \hat q(\theta,a_j)  \cdot  p(\theta,a_j) .
   \label{eq:pay_theta}
   \end{aligned}
\end{equation}

To investigate the influence of this issue on IC property, we conduct comprehensive experiments in Section \ref{re:rq1} to calculate the data-driven IC metric of NMA.
We reserve the discussion about the strictly IC of NMA on complex auction scenarios as an interesting open problem in our future work.

\subsection{List-wise Differentiable Sorting Module}
LDRM can effectively calculate the ranking scores of different ad lists and select the best ad lists for payment.
However, treating allocation and payment outside the model learning (i.e., as an agnostic environment) is in some sense poorly suited for deep learning. That is the processes of allocation and payment (actually the sorting operation) are not natively differentiable. 
\citet{liu2021neural} propose an differentiable sorting engine to solve this problem. But in VCG-based auctions, we sort different ad lists instead of ads, which makes previous solution for different sorting is not suitable in our scenario.
To this end, we propose the list-wise differentiable sorting module, which upgrades the differentiable sorting from point-wise ad ranking to list-wise ad list ranking, making it possible to train the complex auction mechanisms end-to-end.

Given the set $\mathbf{RS} = [\text{RS}(\theta_{1}, \textbf{b}),\text{RS}(\theta_{2}, \textbf{b}),\cdots,\text{RS}(\theta_{N_L}, \textbf{b})]^T$, we define the $\text{argsort}$ operator as the mapping from $N_L$-dimensional real vectors $\mathbf{RS}\in \mathbb{R}^{N_L}$ to the permutation over $N_L$ ad lists, where the permutation matrix $M_{rs}$ is expressed as:
\begin{equation}
  \begin{aligned}
    M_{rs}[j,i]=\begin{cases}
      1 &\text{if}\ i= \text{argsort}(\mathbf{RS}) [j] \\ 
      0 &\text{otherwise} 
      \end{cases},
  \end{aligned}
\end{equation}
where $ M_{rs}[j,i]$ indicates if $\text{RS}(\theta_{i}, \textbf{b})$ is the $j$-th largest ranking score of ad list in $\mathbf{RS}$. The results from \cite{grover2019stochastic} showed the identity:
\begin{equation}
  \begin{aligned}
    M_{rs}[j,i]=\begin{cases}
      1 &\text{if}\ i= \text{argmax}(c_j) \\ 
      0 &\text{otherwise} 
      \end{cases},
      \label{eq:argmax}
  \end{aligned}
\end{equation}
where $c_j = (N_L+1-2j) \ \mathbf{RS} - A_{\text{RS}} \mathbbm{1}$, $A_{\text{RS}}[m,n]$ denotes the absolute pairwise differences of elements and $\mathbbm{1}$ denotes the column vector of all ones. Following previous works \cite{grover2019stochastic,liu2021neural}, we relax the operator argmax in Eq. (\ref{eq:argmax}) as follows:
\begin{equation}
  \begin{aligned}
    \hat M_{rs}[j,:]=\text{softmax}(\frac{c_j}{\tau}),
      \label{eq:softmax}
  \end{aligned}
\end{equation}
where $\tau$ is a temperature parameter. Intuitively, the $j$-th row of $M_{rs}$ can be interpreted as the choice probabilities on all ad lists for getting the $j$-th best ad list. 

We denote the expected revenue of all ad lists calculated by Eq. (\ref{eq:pay_theta}) as $\mathbf{R} = [r(\theta_1),r(\theta_2),...,r(\theta_{N_L})]^T$,
then the end-to-end learning problem of multi-slot auctions can be formulated as minimizing the sum of top-$1$ expected revenue:
\begin{equation}
  \begin{aligned}
    L_{tgt} = - \hat M_{rs}[1,:] \cdot \mathbf{R}.
 \label{eq:tgt}
  \end{aligned}
\end{equation}

\subsection{\!\!Social Welfare Maximization Auxiliary Loss}
As mentioned in Eq. (\ref{eq:tgt}), the ad revenue is only related to the winning ad list, which makes the end-to-end process difficult to learn and leads to a decline of social welfare. Therefore, we design a social welfare maximization auxiliary loss to accelerate the learning process and reduce the decline of social welfare.

Specifically, the social welfare of ad list $\theta$ is defined as:
\begin{equation}
  \begin{aligned}
   SW(\theta, \textbf{b})= \sum_{j=1}^K {b}_j \cdot \hat q(\theta,a_j).
   \end{aligned}
   \label{eq:sw}
\end{equation}
Notice that, we have $v_j=b_j$ in Eq \ref{eq:sw}. Since we follow the theoretical basis of WVCG which is proven IC.

Obviously, the ad list with the maximum social welfare is the result of VCG, where $f_{\mu}(\cdot) = 1$. Here we first form a permutation matrix $M_y$, which is calculated by sorting the social welfare of all ad lists. Then we use row-wise cross-entropy (CE) between the ground-truth permutation matrix and the predicted row-stochastic permutation matrix to construct the Social welfare maximization auxiliary loss:
\begin{equation}
  \begin{aligned}
    L_{ce} = \frac{1}{N_L} \sum_{k=1}^{N_L} \sum_{j=1}^{N_L} \mathbbm{1}(M_y[k,j]=1)\log \hat M_{rs}[k,:].
  \end{aligned}
\end{equation}

The Social welfare maximization auxiliary loss can help the predicted row-stochastic permutation matrix $\hat M_{rs}$ convergence towards maximizing social welfare and effectively reduce the decline of social welfare.

Then the final training loss of NMA is:
\begin{equation}
  \begin{aligned}
    L = L_{tgt} + \alpha_1 L_{ce} + \alpha_2 L_{list},
  \end{aligned}
\end{equation}
where $\alpha_1, \alpha_2$ are coefficients to balance the three losses. We can balance the performance of NMA and keep NMA satisfies the social welfare constraints by adjusting the values of these two parameters.

\section{Experiments}
In this section, we conducted extensive offline experiments on public and industrial datasets
% \footnote{The code is publicly accessible at \url{https://github.com/Lemonace/NMA_code}.}
, and conducted online A/B test on Meituan food delivery platform, with the aim of answering the following research questions:

\noindent $\bullet$ \textbf{RQ1}: Compared with widely used auction mechanisms in the industrial platform, how does NMA perform in terms of platform revenue and social welfare? 

\noindent $\bullet$ \textbf{RQ2}: How do different modules (i.e., CLPM, LDRM, LDSM, SW maximization auxiliary loss) affect the performance of NMA?

\noindent $\bullet$ \textbf{RQ3}: How do different key hyper-parameter settings (i.e., $\alpha_1$, $\alpha_2$) affect the performance of NMA?

\subsection{Experimental Settings}
\subsubsection{Dataset}

\begin{table}[bp]
  \caption{Statistics of the datasets. Avg \#Ads represents the average number of ads per request.\ \ }
  \renewcommand\arraystretch{1.1}
  \centering
  \setlength{\tabcolsep}{4mm}{
  \begin{tabular}{c|cccc}
    \hline
  Dataset & \#Requests  &  \#Users  & Avg \#Ads \\
  \hline
  \hline
  Avito & 53,562,269  &1,324,103   &  10 \\
  Meituan & 230,525,531  &3,201,922  & 20 \\
  \hline
  \end{tabular}
  }
  \label{dataset}
\end{table}

%% 实验结果表
\begin{table*}[tb]
  \renewcommand\arraystretch{1.2}
  \caption{The experimental results of five methods on two datasets. Each result is presented in the form of mean $\pm$ standard deviation (lift percentage). Lift percentage means the improvement of method in this line over the result of NMA.}
  \setlength{\tabcolsep}{3.3mm}{
  \begin{tabular*}{0.95\textwidth}{l|l|cccc}
  \hline
  Dataset &  Model           & CTR               & RPM              & SWPM              & SWMR               \\
  \hline 
  \hline
  \multirow{5}{*}{Avito}
  &  GSP               & 0.0151 $\pm$ 0.0016\ (-8.48\%) & 0.0115 $\pm$ 0.0043\ (-12.21\%) & 0.0133 $\pm$ 0.0025\ (-10.74\%) & 86.93\% \\
  & DNA               & 0.0154 $\pm$ 0.0028\ (-6.67\%) & \textbf{0.0118 $\pm$ 0.0046\ (-9.92\%)} & 0.0131 $\pm$  0.0021\ (-12.08\%) & 85.62\% \\
  & VCG               & \textbf{0.0171 $\pm$ 0.0032\ (+3.64\%)} & \ \ 0.0092 $\pm$ 0.0026\ (-29.77\%) & \textbf{0.0153 $\pm$ 0.0033\ (+2.68\%)} & \textbf{100.00\%} \\
  & WVCG              & 0.0161 $\pm$ 0.0017\ (-2.42\%) & 0.0122 $\pm$ 0.0024\ (-6.87\%)  & 0.0141 $\pm$ 0.0019\ (-5.37\%) & 92.16\% \\
  & \textbf{NMA}       & \textbf{0.0165 $\pm$ 0.0029} & \textbf{0.0131 $\pm$ 0.0025} & \textbf{0.0149 $\pm$ 0.0024} & \textbf{97.39\%}   \\
  \hline 
  \hline 
  \multirow{5}{*}{Meituan} 
  & GSP               & 0.0694 $\pm$ 0.0019\ (-6.97\%) & 0.7534 $\pm$ 0.0041\ (-11.20\%) & 0.8953 $\pm$ 0.0048\ (-6.49\%) & 90.36\% \\
  &DNA               & 0.0721 $\pm$ 0.0016\ (-3.35\%) & \textbf{0.8111 $\pm$ 0.0044\ (-4.40\%)} & 0.8792 $\pm$ 0.0046\ (-8.17\%) & 88.74\% \\
  &VCG               & \textbf{0.0758 $\pm$ 0.0026\ (+1.61\%)} & \ \ 0.7008 $\pm$ 0.0036\ (-17.40\%) & \textbf{0.9908 $\pm$ 0.0032\ (+3.49\%)} & \textbf{100.00\%} \\
  &WVCG              & 0.0739 $\pm$ 0.0038\ (-0.94\%) & 0.8182 $\pm$ 0.0032\ (-3.56\%)  & 0.8881 $\pm$ 0.0037\ (-7.24\%) & 89.63\% \\
  &\textbf{NMA}       & \textbf{0.0746 $\pm$ 0.0017} & \textbf{0.8484 $\pm$ 0.0021} & \textbf{0.9574 $\pm$ 0.0031} & \textbf{96.63\%}   \\
  \hline
  \end{tabular*}}

  \label{tab:result}
\end{table*}

In offline experiments, we provide empirical evidence for the effectiveness of NMA on both public and industrial datasets. The statistics of the two datasets are summarized in Table \ref{dataset}, and we detail the two datasets as follows:
\begin{itemize}[leftmargin=*]
  \item \textbf{Avito}\footnote{\url{ https://www.kaggle.com/c/avito-context-ad-clicks/data}.}. 
  The public Avito dataset comes from a random sample of user search logs from avito.ru. Each search corresponds to a search page with five items, two of which have labels and are regarded as displayed ads, and the other three of which are regarded as organic items.
  For each sample, we construct the set of candidate ads $\mathcal{A}$ with N-2 items clicked by the user and 2 displayed ads in current search, and we generate set $\Theta$ through full permutation algorithm. The public information consists of user ID, search ID and search date. The features of each item consist of item ID, category ID and title. The bid of each ad is independently sampled from uniform distribution between 0.5 and 1.5. 
  Here we use the data from 20150428 to 20150515 as the training set and the data from 20150516 to 20150520 as the testing set to avoid data leakage.
  \item \textbf{Meituan}. 
  The industrial Meituan dataset is collected under GSP auctions on Meituan food delivery platform during April 2022. 
  The sample converted from each request includes two parts: features and labels. The features consist of information of all candidate ad lists and public information.
  These candidate ad lists are generated through full permutation algorithm on the set of displayed ads.
  In one ad list, the information of each ad consists of its bid and its sparse features (e.g., ID, category, brand and so on).
  The labels consist of the final displayed ad list, the ad revenue of this list and the binary click labels of each ad in the list. 
  According to the date of data collection, we divide the dataset into training and test sets with the proportion of 8:2.
\end{itemize}

\subsubsection{Evaluation Metrics}
We construct an offline simulation system, which can ensure that the offline and online performance trends are consistent. Each experiment is repeated 5 times with different random seeds and each result is presented in the form of mean $\pm$ standard. Here
we consider the following metrics in our offline experiments and online A/B tests. For all experiments in this paper, experimental results are normalized to a same scale.
\begin{itemize}[leftmargin=*]
  \item \textbf{Click-Through Rate}. 
  $\text{CTR} = \frac{\sum click}{\sum impression}$.
  \item \textbf{Revenue Per Mille}. 
  $\text{RPM} = \frac{\sum click \times payment}{\sum impression} \times 1000$.
  \item \textbf{Social Welfare Per Mille}.
  $\text{SWPM}= \frac{\sum click \times bid}{\sum impression} \times  1000$.
  \item \textbf{Social Welfare Maximization Ratio}.
  $\text{SWMR}=\frac{\text{SWPM}}{\text{SWPM}^*} \times 100\%$,
\end{itemize}
where $\text{SWPM}^*$ is the SWPM of VCG. We prioritize the revenue indicator while also taking into consideration the social welfare indicator, as it serves as the upper limit for revenue. Despite having the highest social welfare, VCG's revenue is very low. In practical industrial scenarios, social welfare can be reduced within a certain range to ensure maximum income. Notice that, since the real ad revenue is related to the privacy of platform, we convert the absolute value of bid and payment and only show the relative trends of RPM and SWPM.

Apart from above mentioned indicators, we also evaluate the effectiveness of our designed multi-slot auction mechanisms on the property of IC. 
\subsubsection{Hyperparameters}
In both Avito and Meituan dataset, we use the setting of top-$K$ ads displayed (i.e., $K$-slot auctions) in each PV request and $\varepsilon$ is setted as $0.05$ based on business needs.
We tried different hyperparameters in NMA using grid search. Due to space limitations, only the best results are shown in this paper. 
In Avito dataset, $d$ is $8$, $\alpha_1$ is $0.2$, $\tau$ is $1$, $\alpha_2$ is $0.01$, the learning rate is $10^{-3}$, the optimizer is Adam \cite{kingma2014adam}, the batch size is 1,024, the hidden layer sizes of $\text{MLP}_{\text{list}}$ and $\text{MLP}_\mu$ are $(32, 16, 8)$ and $(32, 8, 1)$.
In Meituan dataset, $d$ is $8$, $\tau$ is $0.1$, $\alpha_1$ is $0.4$, $\alpha_2$ is $0.3$, the learning rate is $10^{-3}$, the optimizer is Adam \cite{kingma2014adam} and the batch size is 8,192, the hidden layer sizes of $\text{MLP}_{\text{list}}$ and $\text{MLP}_\mu$ are $(60, 32, 10)$ and $(32, 8, 1)$.
Notice that the number of ads $N$ and the number of organic items $M$ in experiments are truncated for simplicity. So $K$ is $2$, $N$ is $10$, $M$ is $4$ in Avito dataset and $K$ is $4$, $N$ is $20$, $M$ is $50$ in Meituan dataset. But it is obvious that NMA can be extended to scenarios with more ads and organic items.

\subsubsection{Baselines}
We compare NMA with the following four representative auction mechanisms, which are widely used in the industrial ad platform:
\begin{itemize}[leftmargin=*]
  \item \textbf{Generalized Second Price auctions (GSP)} \cite{edelman2007internet}. 
  The ranking score in the classical GSP is simply the bids multiplying $pCTR$, namely effective Cost Per Milles (eCPM). We denote the ranking score of $i$-th ad is $rs_i = bid_i \times pCTR_i$. Its payment is $p_i = rs_i^{-1}(rs_{i+1}(b+1))$, where $rs_{i+1}(b+1)$ is the ranking score of the next highest advertiser and $rs_i^{-1}$ is the inversion function of $r_i(\cdot)$.
  \item \textbf{Deep Neural Auctions (DNA)} \cite{liu2021neural}. 
  DNA is end-to-end neural auction mechanisms based on GSP. DNA can optimize multiple performance metrics using the real feedback from historical auction outcomes.
  In this paper, we uniformly optimize eCPM for both DNA and NMA to ensure fair comparison. But it is worth noting that our experimental conclusions are easily analogous to experiments with multiple performance metrics. 
  
  \item \textbf{Vickrey-Clarke-Groves (VCG)} \cite{varian2014vcg}. VCG with externalities evaluates all allocations using SW. The allocation with the largest SW wins and its payment rule is: each ad in the winning allocation is charged by the loss of SW, which is the difference between the SW of the best allocation without this ad and the SW of the winning allocation.
  
  \item \textbf{Weighted Vickrey-Clarke-Groves (WVCG)} \cite{gatti2015truthful}. WVCG linearly weights the SW of allocation with parameters and solves for the parameters with MAB.
\end{itemize}

%% IC 实验结果
\begin{table}[bp]
  \caption{The experimental results about IC-R of GFP, VCG and NMA on  two datasets.}
  \renewcommand\arraystretch{1.1}
  \centering
  \setlength{\tabcolsep}{6.3mm}{
  \begin{tabular}{c|c|c}
    \hline
  Dataset & Method    & IC-R  \\
  \hline
  \hline
  \multirow{3}{*}{Avito}
  & GFP   & \textbf{0.309 $\pm$ 0.052}  \\
  & VCG  & 0.0 $\pm$ 0.0 \\
  & NMA  & 0.0 $\pm$ 0.0 \\
  \hline
  \hline
  \multirow{3}{*}{Meituan}
  & GFP   &  \textbf{0.195 $\pm$ 0.011}   \\
  & VCG  & 0.0 $\pm$ 0.0  \\
  & NMA  & 0.0 $\pm$ 0.0  \\
  \hline
  \end{tabular}
  }
  \label{tab:result_ic}
\end{table}

\begin{table}[bp]
  \caption{The experimental results about AUC and Logloss of w/ and w/o CLPM on two datasets.}
  \renewcommand\arraystretch{1.1}
  \centering
  \setlength{\tabcolsep}{2.7mm}{
  \begin{tabular}{c|c|cc}
    \hline
  Dataset & pCTR    & AUC  & LogLoss  \\
  \hline
  \hline
  \multirow{2}{*}{Avito}
  &  pCTR by CLPM   & \textbf{0.759 $\pm$ 0.0006}  &  \textbf{0.0481}  \\
  & point-wise pCTR & 0.753 $\pm$ 0.0005  & 0.0483  \\
  \hline
  \hline
  \multirow{2}{*}{Meituan}
  &  pCTR by CLPM   &  \textbf{0.708 $\pm$ 0.0008}  &  \textbf{0.1871}  \\
  & point-wise pCTR & 0.692 $\pm$ 0.0009  & 0.1886  \\
  \hline
  \end{tabular}
  }
  \label{tab:result_auc}
\end{table}

\subsection{Offline Experiments}
\subsubsection{Performance Comparison (RQ1)}
\label{re:rq1}
We construct an offline simulation system, which can ensure that the offline and online performance trends are consistent. Each offline experiment is repeated 5 times with different random seeds and each result is presented in the form of mean $\pm$ standard.
We summarizes the detailed experimental results on public and industrial datasets in Table \ref{tab:result}. 
Compared with representative auction mechanisms, we have the following observations from the experimental results: i) Intuitively, NMA makes great improvements over DNA and GSP in all three metrics on both datasets. One reasonable explanation is that NMA can model global externalities effectively while DNA only model local externalities. ii) Compared with VCG, NMA has a better balance between ad revenue and social welfare, which means NMA achieve higher ad revenue with a small decline of social welfare. It also indicates that VCG maximizes social welfare at the expense of ad revenue, which is unacceptable for most industrial practice. iii) It is obvious that WVCG is overwhelmed by NMA, especially in RPM. This experimental result verifies the powerful expressive ability and efficient end-to-end learning ability of NMA.

In addition, we also apply IC-R \cite{wang2022designing}, which represents the ex-post regret of utility maximizers, to quantify IC of NMA. A larger value of IC-R indicates that an advertiser could get larger utility by manipulate bidding. 
For instance, $0.309$ in Table \ref{tab:result_ic} means advertisers can increase their utilities by about 30.9\% through modifying their bid in GFP auctions.
Specifically, we conduct $2000 \times 10$ tests on Avito and $2000 \times 25$ tests on Meituan each time, where $2000$ is the number of tested auctions and $10$ (or $25$) is the number of ads per auction. For an ad $a_i$, we only replace its bid $b_i$ with $\beta \times b_i$, where $\beta \in \{0.1, 0.3, 0.5,\cdots,1.9\}$ is a multiplicative perturbation factor. All the features and bids of other ads in this auction remains the same. 
Based on above mentioned settings, we random sample data and repeat the experiments $20$ times to observe the IC-R of three auctions.
As is shown in Table \ref{tab:result_ic}, NMA has the same performance with VCG - both are better than Generalized First Price (GFP), which verifies that NMA effectively satisfies IC constraints. 

\subsubsection{Ablation Study (RQ2)}
To verify the impact of different modules (i.e., CLPM, LDRM, LDSM, social welfare maximization auxiliary loss), we study three ablated variants of NMA on Meituan dataset:
\begin{itemize}[leftmargin=*]
  \item  NMA (-CLPM) does not use the context-aware list-wise prediction module and uses original point-wise pCTR to calculate ranking score. For allocations with the same ranking score, we randomly choose one as the winning ad list.
  \item NMA (-LDRM-LDSM) blocks the list-wise deep rank module and the list-wise differentiable sorting module together and uses WVCG to calculate the ranking score and payment of each allocation instead.
  \item NMA (-SW Aux Loss) removes the auxiliary loss for maximizing social welfare, i.e., $\alpha_2=0$.
\end{itemize}

Judging from the experimental results in Table \ref{tab:result_ablation}, we have the following findings: 
i) The variant without CLPM performs worse than NMA. This phenomenon proves that our proposed CLPM can effectively model the global externalities and help NMA to achieve better performance.
ii) The experimental results of NMA(-LDRM-LDSM) are worse than those of NMA in all three metrics. This supports that the upgrade of training can contribute to performance improvement.
iii) The SWPM performance gap between w/ and w/o social welfare maximization auxiliary loss is obvious. This indicates that the auxiliary loss enables us to maximize ad revenue while reducing the decline of social welfare.

Besides, we additionally observe the Area Under Curve (AUC) and Logloss indicators for w/ and w/o CLPM to verify the improvement in CTR prediction by CLPM. As is shown in Table \ref{tab:result_auc}, CLPM improves over the original point-wise pCTR results /emph{w.r.t.} AUC by 0.006 and 0.016 on Avito and Meituan, respectively, which illustrates that CLPM enables effectively modeling of global externalities and improves the accuracy of predictive values.

\begin{table*}[tb]
  \renewcommand\arraystretch{1.2}
  \caption{The experimental results of three different variants of NMA on Meituan. Each result is presented in the form of mean $\pm$ standard deviation (lift percentage). Lift percentage means the improvement of method in this line over the result of NMA.}
  \setlength{\tabcolsep}{4.3mm}{
  \begin{tabular*}{0.95\textwidth}{l|cccc}
  \hline
  Model           & CTR               & RPM              & SWPM              & SWMR               \\
  \hline 
  \hline
   \textbf{NMA}       & \textbf{0.0746 $\pm$ 0.0017} & \textbf{0.8484 $\pm$ 0.0021} & \textbf{0.9574 $\pm$ 0.0031} & \textbf{96.63\%}   \\
  \quad -CLPM        & 0.0689 $\pm$ 0.0018\ (-7.64\%) & 0.7841 $\pm$ 0.0027\ (-7.58\%) & 0.8863 $\pm$ 0.0015\ (-7.43\%) & 89.45\% \\
   \quad -LSRM-LDSM   & 0.0741 $\pm$ 0.0042\ (-0.67\%) & 0.8153 $\pm$ 0.0018\ (-3.90\%) & 0.8987 $\pm$ 0.0029\ (-6.13\%) & 90.70\% \\
   \quad -SW Aux Loss & 0.0744 $\pm$ 0.0034\ (-0.27\%) & 0.8738 $\pm$ 0.0039\ (+2.99\%) & 0.8941 $\pm$ 0.0034\ (-6.61\%) & 90.24\% \\
  \hline
  \end{tabular*}}

  \label{tab:result_ablation}
\end{table*}

\begin{figure*}[btb] 
  \centering 
  \subfigure[{The RPM and SWPM curves of $\alpha_1$.}]{ 
    \centering
    \begin{minipage}{0.43\linewidth}
    \includegraphics[width=\textwidth]{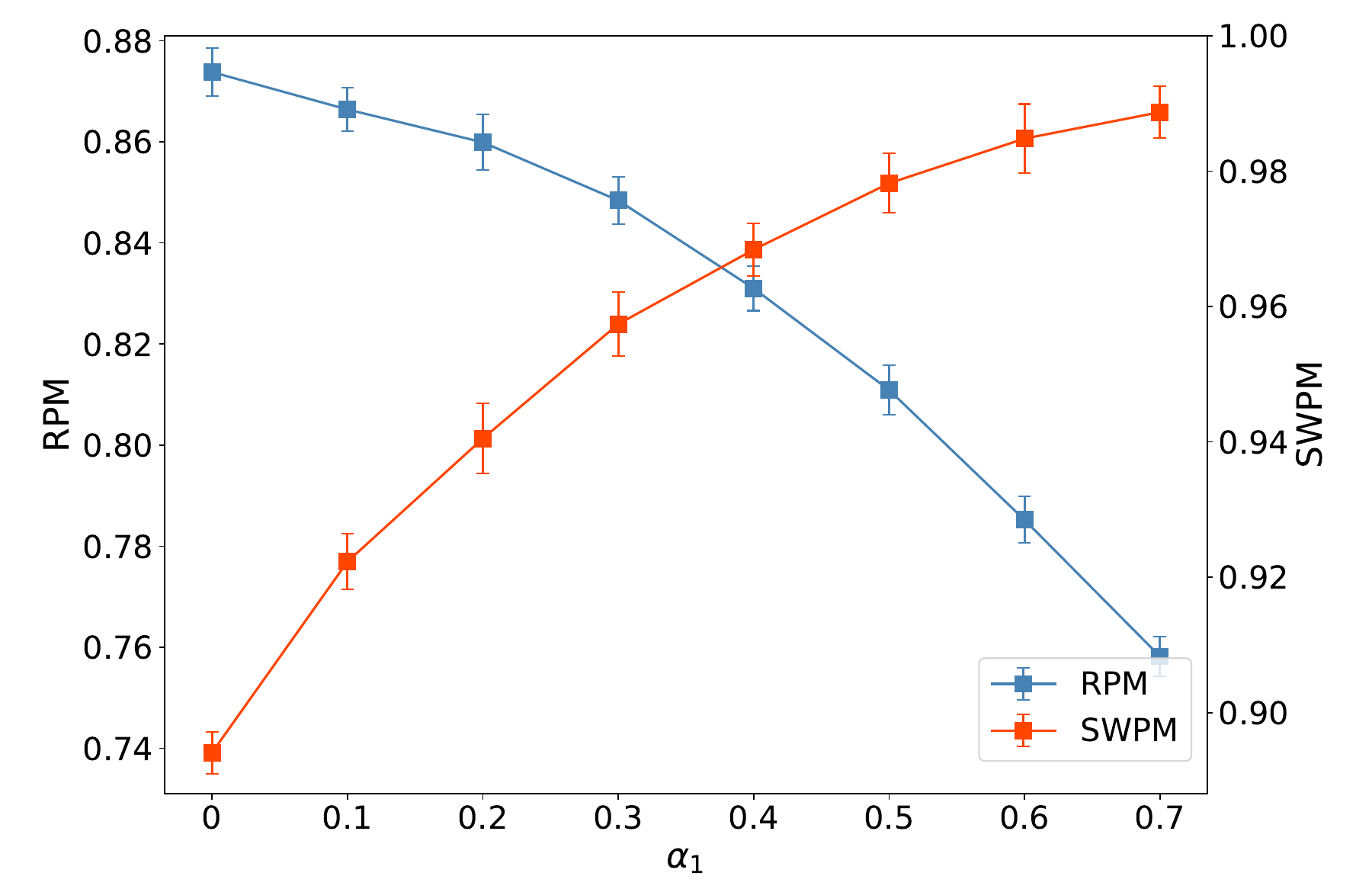} 
    \label{fig:subfig:a1} 
  \end{minipage}%
  } 
  \subfigure[{The RPM and SWPM curves of $\alpha_2$.}]{ 
    \centering
    \begin{minipage}{0.43\linewidth}
    \includegraphics[width=\textwidth]{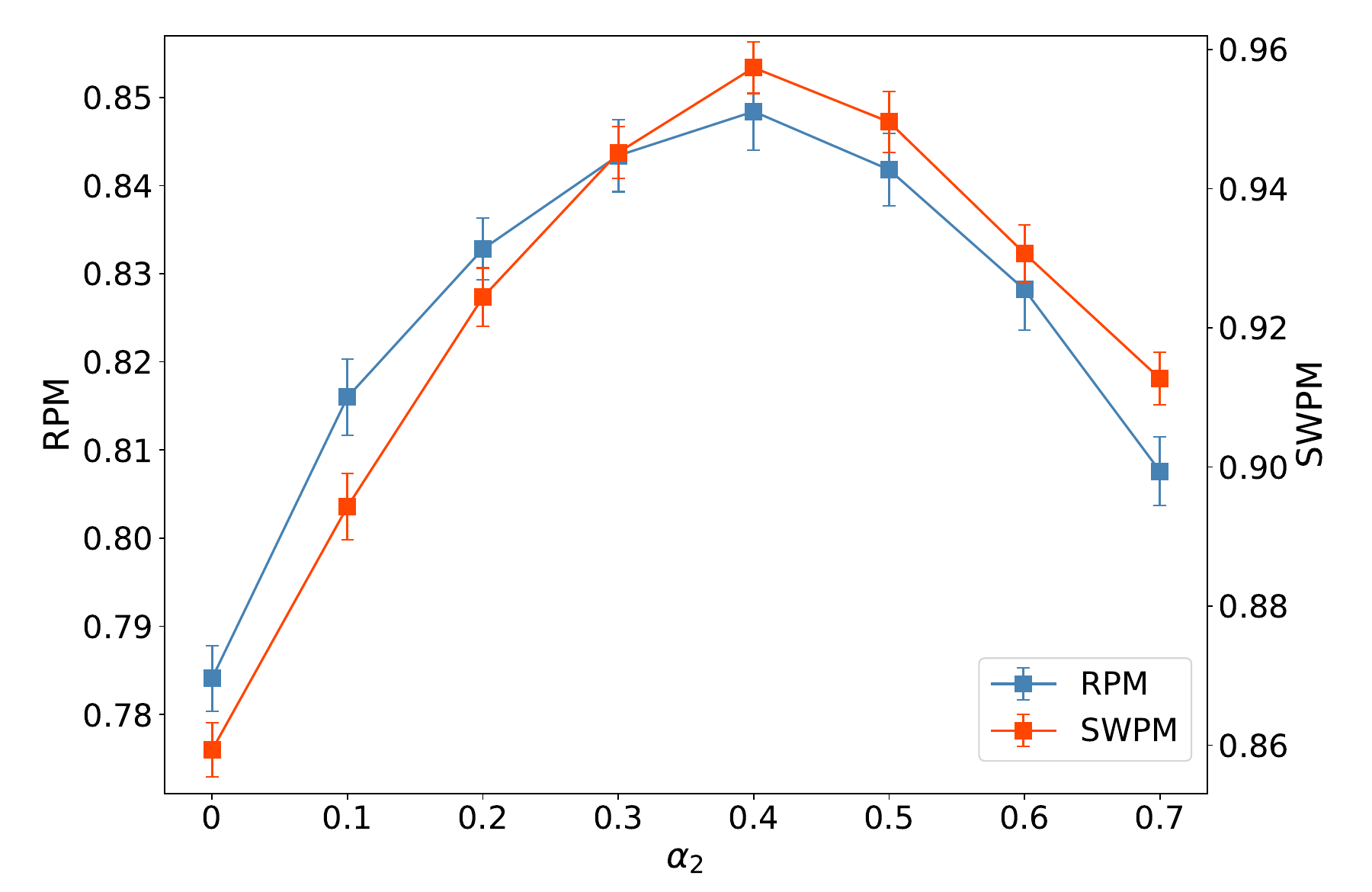} 
    \label{fig:subfig:a2} 
  \end{minipage}%
  } 
  \caption{{The experimental results on the sensitivity of $\alpha_1$ and $\alpha_2$.}
  }
  \label{fig:alpha}
\end{figure*}

\subsubsection{Hyperparameter Analysis (RQ3)}
  We analyze the sensitivity of two hyperparameters: $\alpha_1$, $\alpha_2$.
  Specifically, $\alpha_1$, $\alpha_2$ is the coefficients of the auxiliary loss for social welfare maximization and the auxiliary loss for CLPM, respectively.
  The curves of RPM and SWPM are shown in figure \ref{fig:alpha} and we have the following findings:  
  i) As $\alpha_1$ increases, SWPM of NMA increases but RPM of NMA decreases. This phenomenon shows that our auxiliary loss for social welfare maximization helps NMA achieve a better balance between ad revenue and social welfare. When $\alpha_1$ increases, we pay more attention to social welfare. When $\alpha_1$ decreases, we focus more on ad revenue.
  ii) Increasing $\alpha_2$ within a certain range can improve the performance. But if $\alpha_2$ are too large, it would cause the performance to degrade. One possible explanation is that the auxiliary loss for CLPM can effectively guide the NMA to learn in a target direction within a certain range. But if the weight of  auxiliary task is too large, it may cause the learning direction to be dominated by this task, resulting in a decrease in the overall performance.

\subsection{Online Results}
  We compare NMA with GSP and both auction mechanisms are deployed on Meituan food delivery platform through online A/B test. For candidate ad lists, we use full permutation algorithm to obtain all candidate ad lists online, and the hyperparameters $K$,$N$,$M$ are the same as offline. Notice that, the number of ads $N$ is quite small in our location-based services (LBS) scenario. 
  So we also extend NMA to scenarios with a large number of ads to verify its feasibility, in which we use some list generation algorithms to reduce the time complexity and approximate the generation of candidate ad lists.
  
  We conduct online A/B test with 1\% of whole production traffic from May 20, 2022 to June 20, 2022 (one month). The performance is stable under the long-term observation. As a result, we find that CTR, RPM and SWPM increase by \textbf{6.37\%, 10.88\%} and \textbf{6.22\%} respectively, which demonstrates that NMA can achieve higher platform revenue and social welfare.
  It is worth noting that this increase values are 6.97\%, 11.20\% and 6.49\% in offline experiments. Some possible reasons for this difference in absolute value are the differences in data distribution and small errors in offline evaluation.

\section{Conclusions}
  In this paper, we have proposed the Neural Multi-slot Auctions (NMA) with externalities, towards learning end-to-end auction mechanisms under multi-slot setting for online advertising and obtaining higher revenue with less decline of social welfare. 
  Specifically, we model the global externalities effectively with the context-aware list-wise prediction module. We design a list-wise deep rank module to guarantee IC in end-to-end learning. We also propose a
  social welfare maximization auxiliary loss  to effectively reduce the decline of social welfare while maximizing revenue.
  We have deployed the NMA mechanism on Meituan food delivery platform. 
  The offline experimental results and online A/B test showed that NMA significantly outperformed other existing auction mechanism baselines.
  It is worth considering that the full permutation can be substituted with a list generation module, such as heuristic methods, which can meet industry deployment needs. However, they may have an impact on effectiveness. Therefore, further optimization is necessary to improve work efficiency in the future.
  \balance
\bibliographystyle{ACM-Reference-Format}
\newpage
\bibliography{dca}
\end{document}